\documentclass{article}
\usepackage{arxiv}
\usepackage{times}
\usepackage{latexsym}
\usepackage{url}
\usepackage{graphicx}
\usepackage{hyperref}
\usepackage{longtable}
\usepackage{amsmath}
\usepackage{authblk}
\usepackage[titletoc,title]{appendix}

\usepackage{biblatex} %Imports biblatex package
\usepackage{floatpag}

\usepackage{newfloat}
\DeclareFloatingEnvironment[name={Supplementary Figure},fileext=lsf,listname={List of Supplementary Figures}]{suppfigure}

\usepackage{microtype}

\title{Architectures of Meaning\\ A Systematic Corpus Analysis of NLP Systems}

\author[1,2]{Oskar Wysocki}
\author[1]{Malina Florea}
\author[2]{D\'onal Landers}
\author[1,2,3]{Andr\'e Freitas}
\affil[1]{Department of Computer Science, The University of Manchester}
\affil[2]{digital Experimental Cancer Medicine Team, Cancer Biomarker Centre,\authorcr CRUK Manchester Institute, University of Manchester}
\affil[3]{Idiap Research Institute}

\begin{document}

\date{}
\maketitle

\begin{abstract}
This paper proposes a novel statistical corpus analysis framework targeted towards the interpretation of Natural Language Processing (NLP) architectural patterns at scale. The proposed approach combines saturation-based lexicon construction, statistical corpus analysis methods and graph collocations to induce a synthesis representation of NLP architectural patterns from corpora. The framework is validated in the full corpus of Semeval tasks and demonstrated coherent architectural patterns which can be used to answer architectural questions on a data-driven fashion, providing a systematic mechanism to interpret a largely dynamic and exponentially growing field.
\end{abstract}

\section{Introduction}

Natural Language Processing (NLP) systems have been subjected to a Cambrian explosion of architectural paradigms in the past few years. The scale on the number of contributions and its exponential growth, bring challenges in understanding how NLP architectural patterns evolve and consolidate in different sub-areas and tasks.

This paper aims to provide the methodological support for the interpretation of NLP architectural patterns at scale by applying statistical corpus analysis methods over large-scale NLP corpora. We analyse the use of corpus statistics to compute large-scale collocation patterns jointly with graph visualisation methods as a device to interpret architectural patterns at scale. The proposed methods aims to address questions such as:
\begin{itemize}
    \item What is the complete list of architectural patterns present in NLP?
    \item What are the prevailing architectural patterns (classifiers, layers, regularisation, linguistic resources) for each NLP task?
    \item How these patterns are evolving over time and what are the emerging consolidated/canonical architectural motifs?
    \item What is the architectural variance across different subareas?
\end{itemize}

This paper proposes these specific novel contributions:

\begin{enumerate}
    \item The use of a systematic statistical corpus analysis to interpret architectural features of NLP systems.
    \item The extension of existing corpus statistics methods with a-priori sequence mining to support the observation of architectural motifs.
    \item The proposal of collocation graphs to support the interpretation of more complex architectural motifs.
    \item The validation of the proposed model by the observation of the evolution and consolidation of distinctive architectural patterns.
    \item The publication of a framework for systematic corpus analysis of architectural patterns which can be transported to other AI/NLP corpora.
\end{enumerate}

This paper is organised as follows: Section 2 provides a critical analysis of existing corpus statistics and adapt them to support the proposed architecture mining framework; Section 3 describes the creation of the target corpus, which is followed by Section 4 which describes the construction of the supporting component lexicon. Section 5 provides an empirical analysis of the proposed method for interpreting NLP architectural patterns, which is followed by related work and conclusions.

\section{Statistical Corpus Analysis}

In this section we introduce the corpus statistics measures targeting the extraction of architectural patterns.

\subsection{Relative frequencies}

\begin{figure}[b!]
\centering
\includegraphics[width= 1\textwidth]{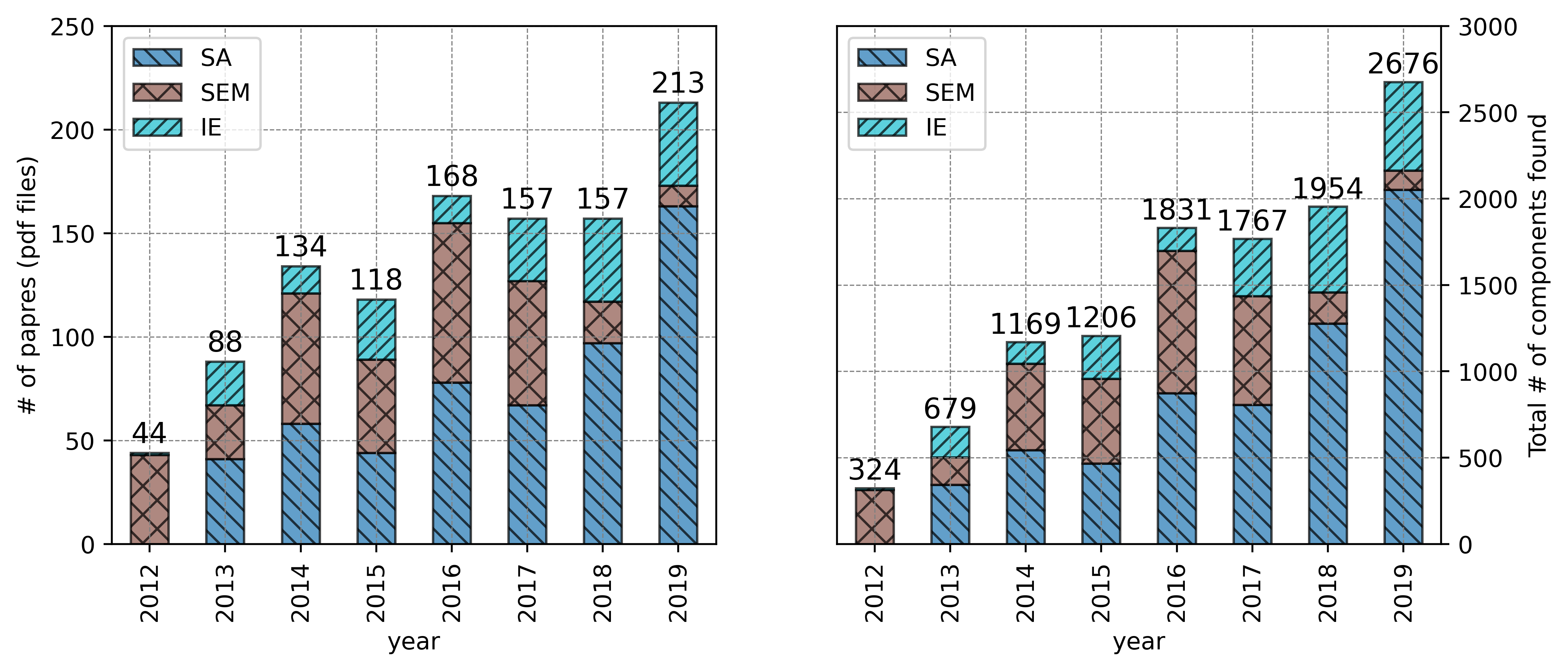}
\caption{ \# of papers with a NLP system description submitted to SemEval (left) and the total \# of components found in these papers in years 2012-2019 (right).}
\label{fig:nb_of_papers}
\end{figure}

The first step in the analysis was to investigate the relative frequency of a single component: $freq_{rel} = \textrm{frequency in given year}/ \textrm{ \# of papers in given year}$.  Both single and multiple appearances of a component mention in a paper count as a binary feature. The advantage of using \textit{relative frequency} instead of \textit{frequency} is to account for a variation in topic popularity along different years (Fig.\ref{fig:nb_of_papers}).

To analyse trends over the years, for each  $freq_{rel}$ data point we fit a linear regression model (LR). We used the LR coefficient \textit{a}, which signifies the slope of the line, and $r^2$ to measure of how well observed values are replicated by the model (Fig.\ref{fig:top10_SA}).%We report the \textit{a} coefficient from the linear regression formula and the corresponding 

\subsection{Collocations}

A \textit{collocation} is defined as a co-occurrence of two or more components in the list of components found in a targeted section of a paper. In order to discover emerging, fading or prevailing collocations over the years, first we propose a pairwise component collocation measure using $G^2$ (the likelihood ratio test statistics), extending it to longer collocations of up to 8 components using the Apriori algorithm \cite{10.5555/645920.672836}. 

In order to define the collocation measure, a co-occurrence matrix was computed for each year, where row names and column names consist of components, and values indicate how many papers two components co-occurred. The most frequent components are more likely to be part of frequent collocations. Thus, in order not to overlook less frequent but significant collocations, we considered 4 statistical measures: \textit{Point-wise Mutual Information} (PMI), \textit{t-score}, likelihood ratio test statistics $G^2$ and \textit{odds ratio} ($\alpha$). PMI tends to strongly favor low-frequency words, whereas \textit{t-score} does just the opposite. $G^2$ appears to be a quasi-standard and was used to rank collocations \cite{gries_2015}. Although $G^2$ is used here as a measure of significance, another measure i.e. $\alpha$ must be used to evaluate whether occurrence of components is positively or negatively associated. $\alpha$ ranges from 0 to $\infty$ , where values $>$1 and $<$1 indicate positive and negative association respectively. Collocations with $\alpha$ $<$ 1 were excluded from the ranking. 

To investigate collocations of 3 or more words, we used the Apriori algorithm, which performs frequent item set mining and association rule learning. A rule is defined as ${X \Rightarrow Y}$,  where $X$ and $Y$ are left and right hand side part of the rule respectively. Both $X$ and $Y$ can be a single component or a set of components. Rules were learned using the \textit{Efficient-Apriori} Python implementation (\cite{efficeintapriori}), with the parameters set on $supp_{min}=0.1$, $conf_{min}=0.3$. This parametrisation expresses that only the components present in at least 10\% of papers in a given year, and that only the rules appearing in at least 30\% of lists containing the left-hand side rule component are considered. There is a variety of measures for the evaluation of association rules \cite{TAN2004293}. We decided to rank the rules using $G^2$, both due to its statistical significance estimation power and for consistency with 2-components collocations comparison. Again, \textit{odds ratio} was used to remove negatively association from the ranking.  
%\[RPF(X \Rightarrow Y) = supp(X \Rightarrow Y) \times conf(X \Rightarrow Y) \]

\[conf(X \Rightarrow Y)= \frac{supp(X \cap Y)}{supp(X)} \]
\[supp(X) = \frac{|{t \in T; X \subset t}|}{|T|} \]

\noindent where  T - set of lists, t - list of components , X - set of components.
For the sake of space, we decided to present the top 10 rules in a given year of SemEval. However components tend to be repeated in rules with similar significance (e.g. \((A,C,E) \Rightarrow (B,D) \) and \((A,C) \Rightarrow (B,D,E) \)). Thus, as the main aim is to draw attention to components rather than to precise rules, each rule is converted into a collocation by merging the components from the $X$ and $Y$ (e.g. \((A,C,E) \Rightarrow (B,D) \) into ($A,B,C,D,E$)). Duplicated collocations are removed, keeping the highest $G^2$. $G^2$  and $\alpha$ are given by:
\begin{multline*}
G^2 = 2* \sum_{i=1}^{4} obs*\log \frac{obs}{exp} = 
2 * ( a_{obs}\log\frac{a_{obs}}{a_{exp}} + b_{obs}\log\frac{b_{obs}}{b_{exp}} +c_{obs}\log\frac{c_{obs}}{c_{exp}}+ 
d_{obs}\log\frac{d_{obs}}{d_{exp}}   )
\end{multline*}
\[\alpha = \frac{ a \times d}{ b \times c}\]
\noindent where $a$, $b$, $c$ and $d$  correspond to the frequencies of collocations $A\&B$, $\overline{A}\&B$, $A\&\overline{B}$ and $\overline{A}\&\overline{B}$  (for 2-components collocations) or $X\cap Y$, $\overline{X}\cap Y$, $X\cap \overline{Y}$ and $\overline{X}\cap \overline{Y}$  (for Apriori rules), \textit{obs} - observed, \textit{exp} - expected.  $a$, $b$, $c$ and $d$ are defined by the contingency table (Tab.\ref{tab:contingancy_table}), which is also used to compute expected values.

\subsection{Collocation graphs}
To compare results of top frequencies, collocations and their changes over the years, we used \textit{network graphs} \cite{10.1371/journal.pone.0098679} as a joined collocation and visualisation device. Graphs were plotted using \textit{Gephi} \cite{ICWSM09154}. Layouts were first computed using the \textit{ForceAtlas 2} algorithm, and subsequent \textit{Noverlap} (to prevent overlapping nodes) and label adjustment were applied. Each node in a graph corresponds to a component, and each edge to a collocation. Collocation's $G^2$  was used as a weight in {ForceAtlas 2}. Only the top 100 (ranked by $G^2$) collocations were included. It resulted in a constant n\# number of edges, but a varying \# of nodes, which additionally informs about diversity in a given sample.

\begin{longtable}[c]{c|ccc}
\caption{Contingency table used for the calculation of $G^2$ and $\alpha$ for a collocation $A\&B$. A and B are components.}
\label{tab:contingancy_table}\\
\textbf{} & \textbf{$A$} & \textbf{$\overline{A}$} &  \\\hline
\endfirsthead
\multicolumn{4}{c}%
{{\bfseries Table \thetable\ continued from previous page}} \\
\textbf{Collocation} & \textbf{A} & \textbf{A} &  \\
\endhead
\textbf{$B$} & a & b & a+b \\
\textbf{$\overline{B}$} & c & d & c+d \\
\textbf{Totals} & a+c & b+d & a+b+c+d
\end{longtable}

\section{Creating the Target Corpus}
In order to automatically and systematically analyse NLP systems (in terms of components, representations and features) we curated 2 Corpora, named \textbf{SemEval Corpus} and \textbf{ACL Corpus}. Both are based on scientific research publications from a top NLP conference and the major task-based challenge workshop.

\noindent \textbf{SemEval Corpus}: All papers from the SemEval Proceedings from 2012 to 2019 were collected, corresponding to 1341 papers. Each paper was converted into a set of text blocks (using PyMuPDF), which were further mapped into labeled sections (e.g. Abstract, Introduction, References). Data cleaning techniques used for each publication included: i) used a rule-based method to avoid repetitions, short blocks and other data modalities, including page numbers or values from tables; ii) removed citations (e.g. 'Johnson et.al 2015'), punctuation marks, numbers (excluding those in names, e.g. w2v) and stop-words; iii) Keeping all paragraphs, removing the “References” sections; iv) the resulting text was tokenised and converted into a list of sentences. The SemEval Corpus consists of a set of 152000 sentences.

\noindent \textbf{ACL Corpus}: All papers from ACL Proceedings from 2010 to 2019 were collected, corresponding to 2265 papers files in total. Using the same steps as for SemEval Corpus, we obtained the ACL Corpus, which was a set of 480000 sentences. Both corpora are available at the project repository (anonymised link). The Semeval corpus will be used to evaluated the proposed approach, while the ACL corpus is used to build the component extractor.

\section{Target terms \& Component lexicon}

%This section explains how we created and used this NLP lexicon along with a collection of NLP publications in order to systematically analyse architectural trends and patterns.  
A critical component of the proposed framework is the creation of a comprehensive lexicon which describes the architectural components including types of classifiers, feature extraction methods, feature representations, gold-standards, data pre-processing methods, among others. 

We developed a \textit{saturation-based method} which coordinated iterative cycles of manual and automatic extraction of components from NLP literature until a plateau in the number of new lexical items was observed (stop point). This method used the iterative application of three steps: 

\noindent \textbf{Seed lexicon}: Building a seed lexicon by the manual curation 2 Machine Learning technical textbooks and 6 comprehensive surveys focused on Sentiment Analysis and Question Answering systems (references described in the Appendix). This list was later enriched with the support of a senior domain expert in NLP. All terms were manually extracted and normalised to a canonical form, adding 322 terms to the lexicon.

\noindent \textbf{Component classifier}: Construction of a Naive-bayes classifier using morphological-level features and context words, in order to determine whether a token is part of a component name or not. The classifier was built over the ACL corpus. In the first cycle the classifier reported 3000 new distinct terms, where 192 of them proved to be true new positive cases. 
%65 were further eliminated for various reasons, and only 25 remaining were added. 5 of them were new component names (e.g. ELMo, Gradient Boosting Machine) and 20 of them were variation of existing entries (e.g. BayesNet for Bayesian Network). Considering the small number of new terms found (5) compared to the initial number of terms (322), we concluded automatic component extraction as 'complete'. 
%(for example we eliminated BiLSTM-GloVe, since it was a combination of 2 existing lexicon terms, BiLSTM and GloVe).

\noindent \textbf{Gold standard}: Manually annotated over 100 randomly sampled papers from the ACL corpus (annotated component mentions). This supported the evaluation of the performance of the algorithm for automatic component extraction (described later in the text) and a post-hoc update of the lexicon with 5 additional items.  %Interestingly enough, the difficulty of covering all component names does not necessarily come from finding new components, but rather finding all similar names used for those components. For example our lexicon contains 11 similar names for the CNN component. These 100 papers was also used

The final lexicon consists of 151 unique canonical components (entities), which are represented by one or more lexical expressions entities, e.g. CNN: cnn, convnet, convolutional neural net, among others, resulting in the total of 519 lexical items. The final lexicon is available in the Appendix.

\textbf{Components extraction}
We clustered all SemEval papers from 2012 to 2019 tasks into 6 macro-categories: Sentiment Analysis (SA), Semantic Analysis (SEM), Information Extraction (IE), Question Answering (QA), Machine Translation (MT) and Other (OT). 74 (77\%) tasks out of the total 96 are tasks SA, SEM or IE: 21, 34 and 19 respectively (summary in the Appendix). 92\% ($1750/1893$) of participating teams and 82\% ($1079/1314$) of submitted papers (with a description of competing system) are associated with SA, SEM and IE tasks. Thus, further analysis includes only SA, SEM and IE related papers, considered as the core of SemEval campaigns. We collected all contributions papers (excluding task description papers). %A procedure in the following paragraph is referred to as \textit{component extraction algorithm} later in the paper. 

From each file, text content was extracted using the PyMuPDF python package. Only the sections related to the description of approaches and systems were kept, filtering by a lexicon of section headings. Text was cleaned and normalised using heuristic rules. Both the headings and the preprocessing rules are fully described in the Appendix. The lexicon maps different lexical forms to a single canonical entity and for a single paper the presence of a component defines a binary features (whether the component is present or not).

\section{Results and discussion}

\subsection{Lexicon evaluation}
 
In order to evaluate our algorithm for automatic component extraction, we developed a Gold Standard for the extraction of NLP architectural components (consisting of 979 examples). It consists of a list of components, one for target sampled paper (out of 100). The evaluation of the extraction method reached a recall=0.884 (std=0.206) and a precision=0.670 (0.227). %This demonstrates a bias towards Only 8 (out of 979) false positives were.    

\subsection{Statistics for SemEval}
There is an increasing trend in the number of papers submitted to SemEval, which correlates with higher total \# of components found in these papers in a given year (Fig.\ref{fig:nb_of_papers}). Note, that the proportion of papers associated with SA, SEM and IE changes, as well as proportion of \# of components. SA gains popularity as SEM decreases, what may be one of the factor in evaluating the shift from one group of components to another over the years.

Over the years we observe an increase in the average number of components found in a paper (Fig.\ref{fig:boxplot_ax4}a), which may have several causes. First, novel systems simply are built in a more sophisticated manner which can entail more components. Researchers experiment more with the architecture and use different components for specific subtasks, taking advantage of variety of available methods developed over the years. Second, the lexicon may be biased in terms of number of unique components towards novel algorithms. For example, novel systems based on BiLSTM are likely to be collocated with at least several other components (related to neural networks, e.g. dense layer, softmax etc), whereas ensemble of a rule-based and SVM model counts only as '3 components'. Thus, we report only a trend of increasing architectural diversity, leaving a detailed analysis of the variability of the components as future work. %Lastly, one could point out, that the later the publication, the higher number of baselines and related works that it can refer to. Thus the noise in extracted component lists is likely to be higher. Yet, as we presented in 5.1, our algorithm has high precision and this argument can be neglected.

For example, novel systems based on BiLSTM are likely to be collocated with at least several other components, whereas ensemble of a rule-based and SVM model counts only as '3 components', while being possibly as complex as the former.

\subsection{Relative frequency}

\begin{figure*}[t!]
\centering
\includegraphics[width= 1\textwidth]{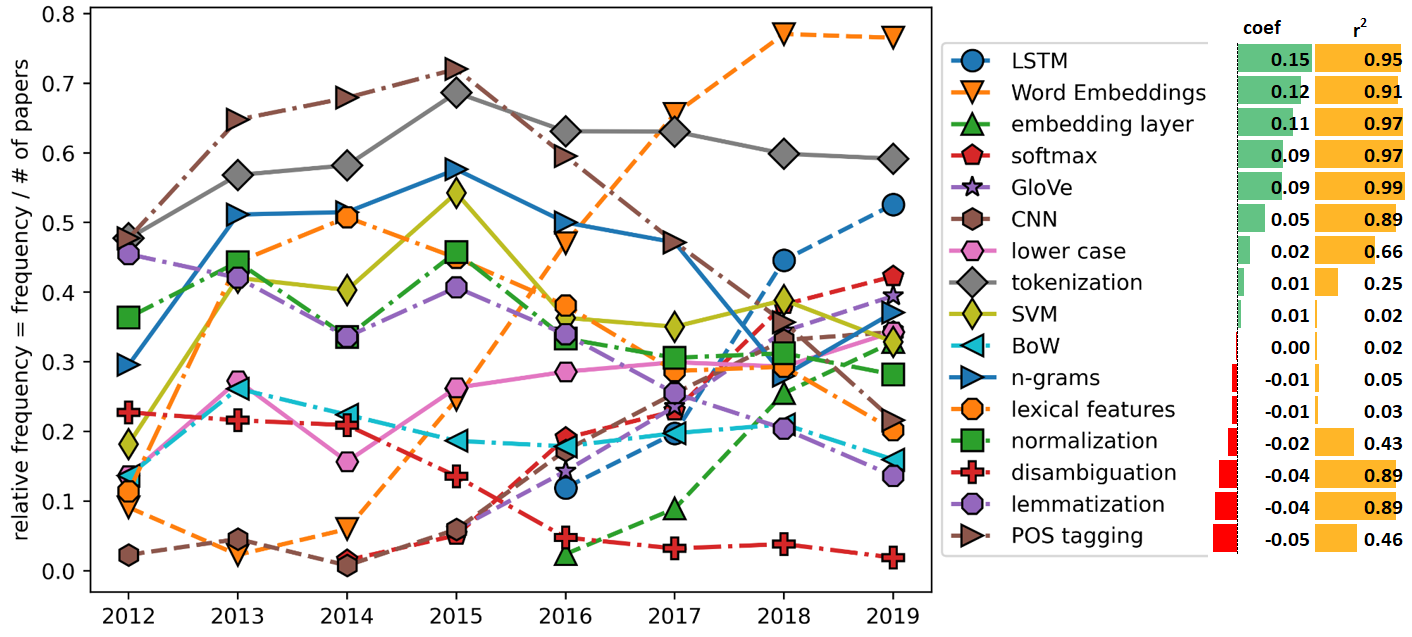}
\caption{The most frequent components in 2013 (dashdot line), in 2019 (dashed line), or both in 2013 and 2019 (solid line). \textit{Coef} is the slope of the regression model for each component and signifies increase or decrease in frequency over years.}
\label{fig:top10_SA}
\end{figure*}

We listed the top 10 most frequently mentioned architectural components in 2013 and 2019. This is depicted in Fig.\ref{fig:top10_SA}. As four of the components ('SVM', 'lower case', 'n-grams' and 'tokenization') were in both lists, we analysed 16 components: 'POS tagging', 'lexical features', 'normalization', 'lemmatization', 'BoW', 'disambiguation' (for 2013, dashdot lines), and 'Word Embeddings', 'LSTM', 'softmax', GloVe', 'CNN', 'embedding layer' (for 2019, dashed lines). These 4 mutual components (solid lines) show longevity, as their $freq_{rel}$ do not significantly change over the years, i.e. coefficients of linear regression fits stay in the $(-0.02, 0.02)$ range. Similarly, 'BoW' also maintains rather constant use ($coef=0.0$). Highest drops are apparent for 'lemmatization' and 'disambiguation': $-0.05$ and $-0.04$ respectively. Although 'POS tagging' $coef=-0.05$ , its $r^2=0.46$ suggests that the LR fit is not perfect, and in fact we observe a frequency increase along the 2012-2015 period, and then a drop until 2019. A significant popularity boost can be observed for 'LSTM', 'Word Embeddings', 'embedding layer', 'softmax' and 'GloVe'. They all have high positive $coef$ values and $r^2 > 0.9$ . A list of top 100 components in 2019, with frequencies, $coef$, $r^2$ is presented in the Appendix.

\subsection{Collocations}
\begin{figure}
\centering
\includegraphics[width= 1\textwidth]{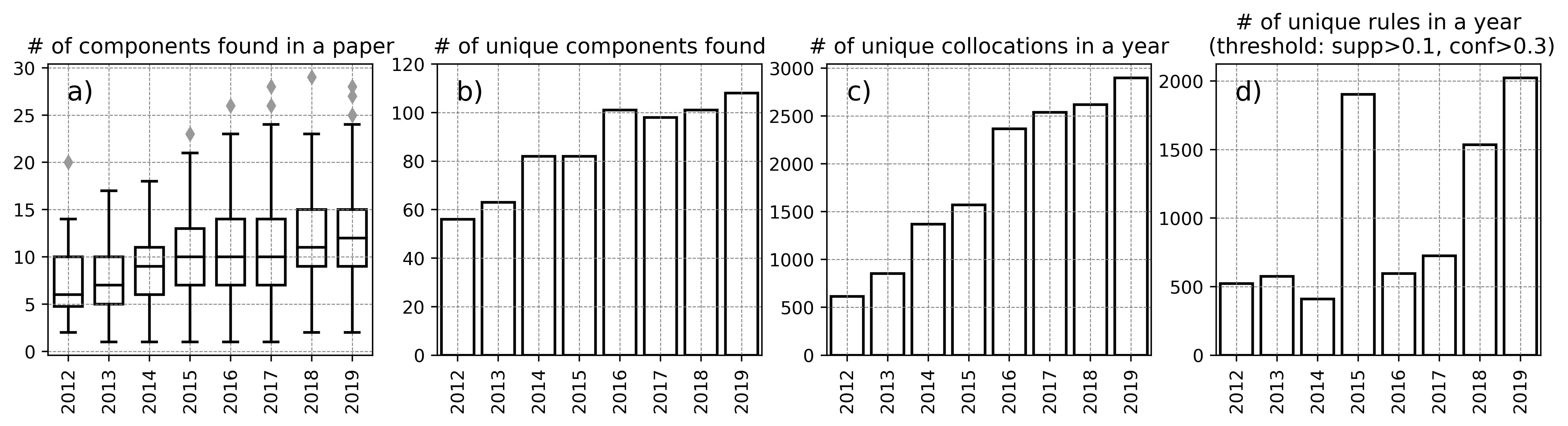}
\caption{Statistics for SemEval 2012-2019: a) \# of components found in a single paper; b) \# of unique  components; c) \# of unique 2-component collocations d) \# of unique rules found by Apriori algorithm.}
\label{fig:boxplot_ax4}
\end{figure}

Similarly to the trend mentioned in 5.2, we observe a higher number of unique collocations found over the years (Fig.\ref{fig:boxplot_ax4}c). This is expected as in general $\# components \sim \# collocations$. A relevant pattern is depicted in Fig.\ref{fig:boxplot_ax4}d, where the \# of unique rules in 2015 is significantly higher than in 2016-2018, not matching trends from former plots. Furthermore, despite higher \# of collocations in 2014 than 2012-2013, again, the \# of unique rules is lower.

Based on Fig.\ref{fig:nb_of_papers} and Fig.\ref{fig:boxplot_ax4} a comparison between consecutive years can be performed. In 2014 more publications were submitted than in 2015 (134 vs. 118), but the \# of components found is almost equal (1169 vs. 1206). This corresponds to the lower \# of components found in a single paper on average (Fig.\ref{fig:boxplot_ax4}a). At the same time, the \# of unique components are equal (both 82, Fig.\ref{fig:boxplot_ax4}b). The higher average \# of components in a paper induces more possible collocations, and this is confirmed in Fig.\ref{fig:boxplot_ax4}c (1370 vs 1570 2-element collocations). What draws the attention, is a prominent difference in the \# of unique rules (409 vs 1902, Fig.\ref{fig:boxplot_ax4}d). To explain it, we need to take a closer look into the Apriori algorithm and components distribution over the papers. Let $N$ be the total \# of components found, $u$ the \# of unique components, $n$ the \# of components in a paper and $m$ the \# of papers, assuming that $N/m<u$. Considering two extreme situations, having  i) minimal and ii) maximal possible total \# of collocations $C$. In i) all components are evenly distributed among papers. Thus, each paper contains $n=N/m$ components. The \# of collocation in each paper equals $n!$ , and $C=m\cdot n!=m\cdot (N/m)!$. In ii) we have $x$ papers containing $u$ components, and $(m-x)$ with only one component, where $x = (N-m)/(u-1)$. Then $C=x \cdot  u! = (N-m)/(u-1)\cdot  u!$. To sum up, in i) $C \sim (N/m)!$ and in ii)  $C\sim u!$  and $N/m \ll u$ (e.g. for year 2014 $N/m=1196/134 \cong  9 \ll u=82$). Thus, the more uneven distribution of the components, the more collocations are expected. The simple intuition behind this is that one long list generates more collocations than several short lists. The same would apply to the rules generated by the Apriori algorithm, if not the thresholds $supp>0.1$ and $conf>0.3$. If in the set of papers there are only several papers with a high \# of components, and a rule belongs to a large group of rules generated from such paper, it is likely that the rule does not pass the $supp$ threshold. Thus, the final \# of unique rules compared with \# of collocations reveals the components' distribution in the papers. High \# in 2015 (Fig.\ref{fig:boxplot_ax4}d) signifies higher variation of frequent collocations, comparing to 2014 (note, that \# of unique components in 2014 and 2015 are $ \cong $ ). In summary, there are many component-triples (or quadruples, up to octuples) consistently occurring in the papers (i.e. at least in 10\% of papers). Similarly, in SemEval 2016-2018 less rules pass the threshold than in 2015, despite the higher \# of components. It again signifies a higher variation of rules in 2015. Note, this reflects a  general comparative analysis of components and collocations frequency, rather than a significance evaluation.

\begin{figure*}[!]
\centering
\includegraphics[width= 1\textwidth]{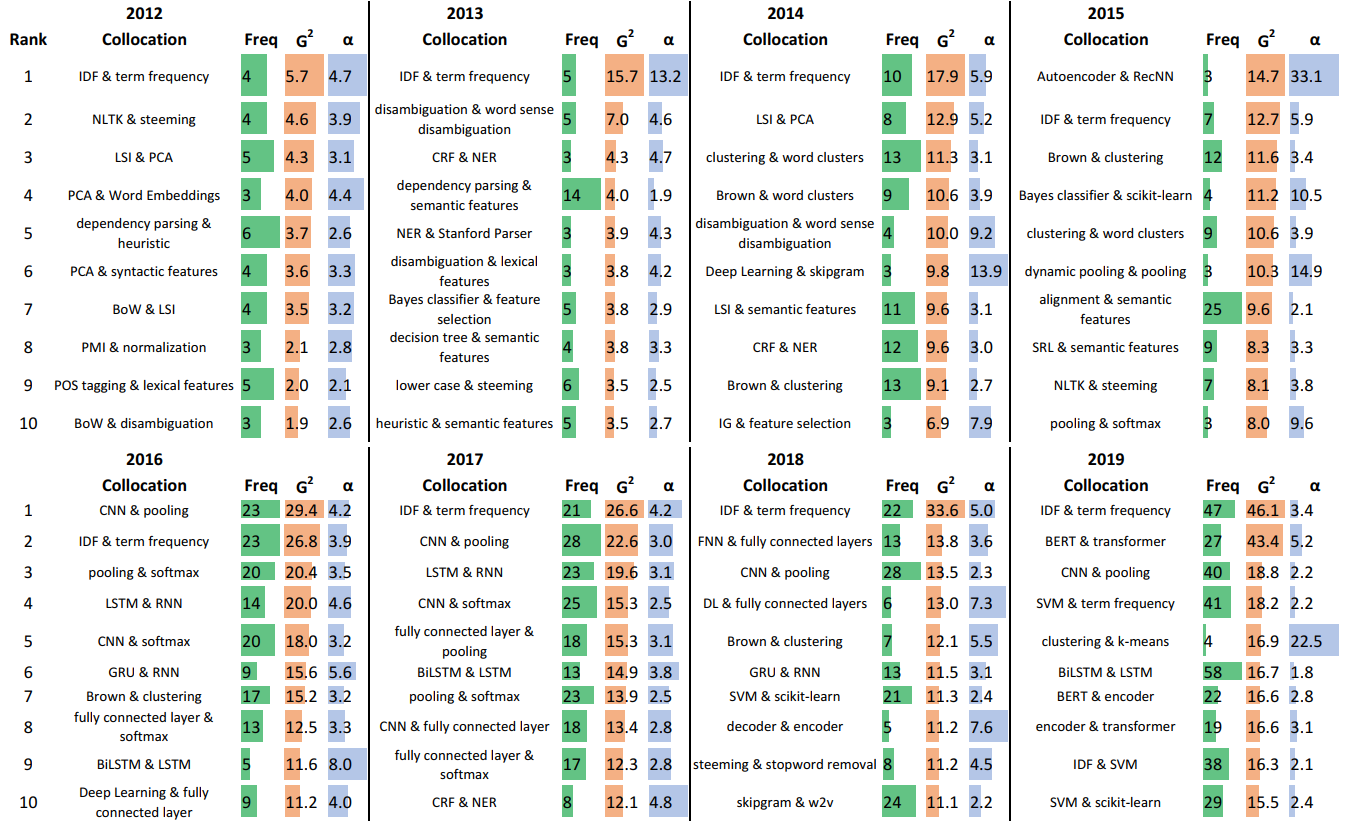}
\caption{Top 10 two-component collocations in SemEval (SA, SEM and IE tasks) for years 2012-2019 ranked by $G^2$ (only positive association: \textit{odds ratio}$\alpha>1$).}
\label{fig:top10_collocations_SA}
\end{figure*}

An analysis of two-element collocations is summarized in Fig.\ref{fig:top10_collocations_SA}, where the top 10 collocations for years 2012-2019 ranked by $G^2$ are depicted, together with corresponding frequencies and odds ratios. A high significance of some collocations is expected in advance. Firstly, for hyponym-hypernym pairs, e.g. IDF \& term frequency (which is the top collocation 6 times), clustering \& k-means, Brown \& clustering or BiLSTM \& LSTM the expectation of high significance is confirmed. Secondly, nearly inseparable pairs due to the model architecture, like CNN \& pooling, pooling \& softmax or decoder \& encoder are also present in the summary. Finally, less expected but still significant collocations show variety and distinguishable trends over the years. From 2016, most of the top collocations consist of deep learning related components, however some of the older methods (like SVM \& term frequency, skipgram \& w2v) seem to maintain their longevity and relevance. Before 2016, a higher variation in the top10 lists is observed, as the there is no obvious choice for best architecture among participating teams. Some of more notable collocations are: CRF \& NER (Conditional random field \& Named-entity recognition, in 2013, 2014, 2017), LSI (Latent semantic indexing) collocations: \& PCA, \& BoW, \& semantic features, among others.

\begin{figure*}[!]
\centering
\includegraphics[width= 1\textwidth]{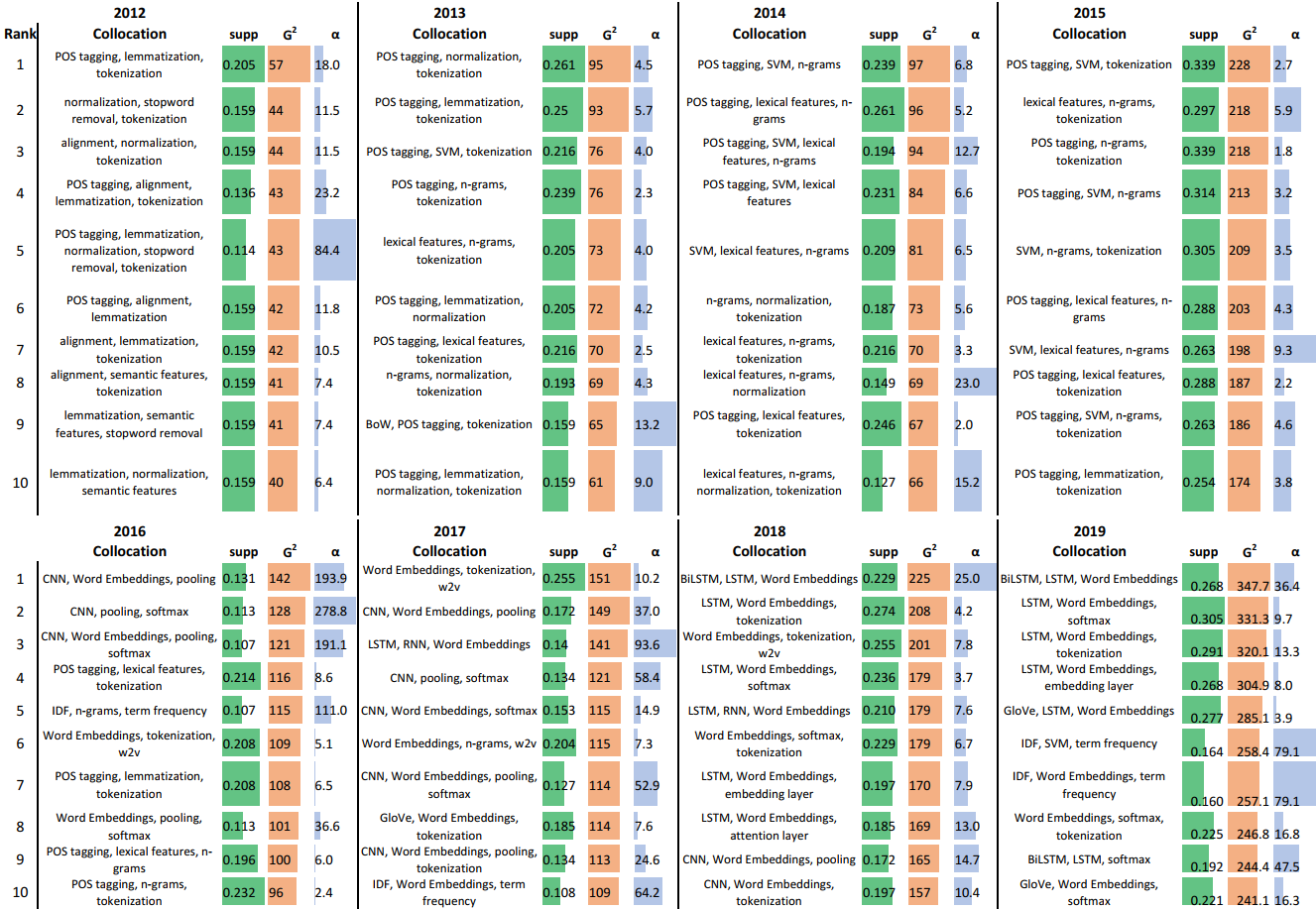}
\caption{Top 10 collocations in SemEval (SA, SEM and IE tasks) derived from rules found using Apriori for years 2012-2019, ranked by $G^2$ (only positive association: \textit{odds ratio }$\alpha >1$).}
\label{fig:top10_apriori_SA}
\end{figure*}

Investigating collocations of 3 or more, we summarized top 10 sets of components for each year, derived from association rules in Fig.\ref{fig:top10_apriori_SA}. In the last 3 years most significant collocations consist of 'LSTM' and/or 'Word Embeddings', with also high visibility of 'CNN' and/or 'tokenization'. In years 2012-2015 various collocations of 'POS tagging' and 'tokenization' are noticeable, which interestingly corresponds to $freq_{rel}$ but not to collocations from Fig.\ref{fig:top10_collocations_SA}. Apart from the mentioned above, some more of the recurring collocations in this summary are: deep learning components ('softmax', 'pooling'), 'n-gram', 'lemmatization' and 'IDF' collocations.

\subsection{Network graphs}
\begin{figure*}%[t!]
\centering
\thisfloatpagestyle{empty}
\includegraphics[width= 1\textwidth]{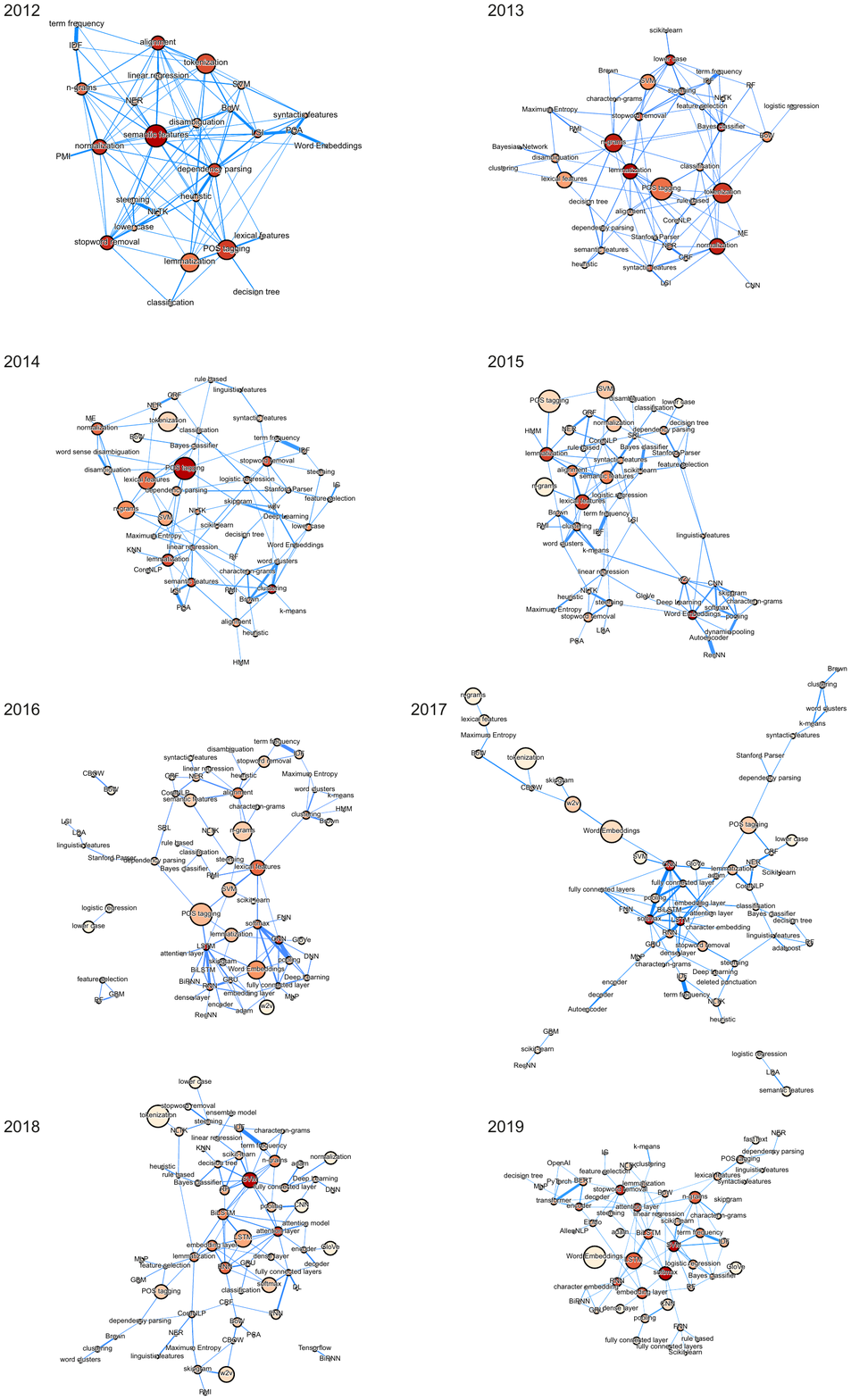}
\caption{Network graphs built using the top 100 collocations from SA, SEM and IE tasks in SemEval from 2012 to 2019. Node size is proportional to the frequency of the component, the red color intensity to node degree, edge width to $G^2$ of the collocation. For full resolution see the digital version of the paper.}
\label{fig:graphs_SA}
\end{figure*}

Summaries discussed above (of 2 and $ \geq 3$  components collocations) are limited to 10 rows. In order to deliver more specific and concise insights about frequencies and collocations, we visualised the top 100 collocations and presented as network graphs in Fig.\ref{fig:graphs_SA}. Note, that most of the observations mentioned in previous paragraphs are confirmed by graphs. Additionally, node size $\sim $ frequency, red color intensity $\sim $node degree (\# of edges connected to the node) and edge width $\sim G^2$ simultaneously inform about the component popularity, the density and significance of connections. Each graph consists of 100 edges, but various \# of nodes.

The network graph for SemEval 2012 is relatively dense, which reflects the fact that in this year only 56 unique components and 613 unique collocations were found. It consist of only 28 nodes, and the high frequency of a component corresponds to the high degree of the node, i.e. 'Semantic features', 'POS tagging', 'tokenization'. Significant edges (collocations) can be found both between high/low frequency/degree nodes. Note, that in 2012 only the SEM task type is reported. In 2013, the graph consists of 41 nodes and central roles play 'POS tagging', 'n-grams', 'lemmatization' and 'tokenization'. 'SVM', 'normalization' and 'lexical features' also appear to be highly relevant representational components. In 2014, again there is an increase of variety (51 nodes). Still, 'POS tagging', 'lexical features', 'n-grams', 'tokenization' are the most noticeable, followed by 'SVM', 'lemmatization' and 'stopword removal'. We observe a growing number of less frequent components.  

In the SemEval 2015 graph (53 nodes), we observe a noticeable split between architectural patterns. In the first cluster, we have again 'lexical' and 'semantic features', 'POS tagging', 'lemmatization', 'n-grams' and 'SVM'. In the other, 'Word Embeddings' with connections to 'w2v', 'CNN', 'softmax' etc. are present, this represents a transition point in the architectural patterns in the direction of neural representation.   

In 2016, we observe a substantial increase in \# of unique components and collocations found, as well as in total \# of components, likewise the variability in the graph: \# of nodes increased to 67. 'Word Embedding' as a center of neural network related nodes, draws the attention to high interest of deep learning and emergence of LSTM usage. However, canonical NLP components (features and pre-processing steps) are still highly present: 'POS tagging', 'lexical features', 'n-grams'  or 'lemmatization' as well as 'SVM'.  

For 2017, the center of the graph is populated by deep learning components, i.e. 'LSTM', 'CNN', 'softmax'. However, $ \geq 3$  collocations from Fig.\ref{fig:top10_apriori_SA} are represented by the upper left branch of the graph, with 'Word Embedding', 'tokenization' and 'w2v'. In 2018 we observe the prevalence of sequence-based DL models ('LSTM', 'RNN' at the center). 'Tokenization' and 'POS tagging' are still frequent. A noticeable change is that we observe 'SVM' with other ML methods like 'RF', 'Bayes classifier', 'decision tree' to constitute the upper part of the graph. In the lower part the connection between 'w2v' and 'skipgram' can be highlighted.

Finally, the SemEval 2019, which is dominated by deep learning, is represented by a graph which is denser than the previous 3 editions (56 nodes, 11 and 8 less than in 2017 and 2018). Apart from the most frequent 'Word Embeddings', which is interestingly connected only to 'LSTM', there are still some old ML components like 'SVM', 'logistic regression' or 'Bayes classifier'. Upper right branch consists of 'POS tagging', lexical' and 'semantic features'. Upper left part highlights the usage of 'BERT' and 'ELMO'. Similarly to 2018, 'attention layer' and 'n-grams' are close to the center. Note, that for the 2019 graph we mainly report components associated with the SA task. 

In summary, graphs illustrate broader collocation patterns ($ \geq 3$) (Fig.\ref{fig:top10_apriori_SA}). They provide a natural normalisation mechanism for putting high-frequency canonical components into context (although term frequency and IDF appear to be very significant in 2-component collocations, they weaken in $ \geq 3$  collocations, and play rather marginal role when analysing the graphs). Moreover they provide a natural way to observe the emergence of clusters of architectural components.

Network collocation graphs built over collocation analysis can support the interpretation of architectural patterns at scale. The analysis confirmed architectural patterns coherent with the major trends on the evolution of the field. The proposed method excelled at inducing functional clusters. Among others, one observable pattern is on the longevity and the cumulative aspect of some architectural motifs, which tend to consolidate and compose with novel functional clusters (e.g. the emerging and recurring architectural theme: POS, lemma $->$ embeddings $->$ Sequence DL architectures $->$ attention $->$ transformers).

The reader is referred to a task-based detailed results for SA, SEM and IE see \hyperref[sec:SuppMat]{Supplementary Material}.

\section{Related work}
%the field is dynamic
%continues update
%systematic
%granular type of survey
%volume
%dynamic

To the best of our knowledge this is the first time that corpus analysis methods were developed to perform a systematic analysis of architectural patterns in NLP and AI or applied for a systematic survey in NLP. In the space of corpus linguistics, Gries \cite{Gries2010},\cite{gries_2015} provides a comprehensive reference for statistical corpus analysis, which we build upon, extending it with a saturation-based lexicon construction methods and with association rule and graph collocation mining. Other comprehensive studies presenting a perspective on corpus linguistics are provided by Hall \cite{hall-etal-2008-studying}, Brezina \cite{brezina_2018}, Mohammad \cite{mohammad2019state}\cite{mohammad-2020-nlp} and Bollmann \cite{bollmann-elliott-2020-forgetting}. On the systematic literature analysis, Wysocki \cite{wysocki2020semeval} describes a comprehensive critical structured survey of the Semeval campaign, focusing on the distribution of task types and their impact. Methodologically, these works emphasise corpus statistics to some extend but the analyses do not provide a granular architectural perspective. 

Surveys targeted on specific task types such as Question Answering  \cite{bouziane2015question} or Sentiment Analysis \cite{medhat2014sentiment} or a specific architectural paradigm \cite{zhang2018deep} are abundant but most commonly they do not apply any statistical analysis method and focus on smaller scale text interpretation (in contrast to a corpus analysis method which can scale up to large reference bases). 

\section{Conclusions}

This paper proposed a novel statistical corpus analysis framework targeted towards the interpretation of NLP architectural patterns at scale. The framework combines a saturation-based lexicon construction, statistical collocation methods and graph collocations to derive an aggregate representation of architectural patterns. The framework was validated in the context of the corpus analysis of Semeval tasks and demonstrated consistent architectural patterns which can be used to address questions on the evolution of NLP architectures in a data-driven manner.

%\bibliography{eacl2021}
%\bibliographystyle{acl_natbib}

\printbibliography %Prints bibliography
\newpage

\section{Supplementary Material}
\label{sec:SuppMat}
%\appendix
\counterwithin{figure}{section}

\begin{suppfigure}[h!]
\centering
\includegraphics[width= 1\textwidth]{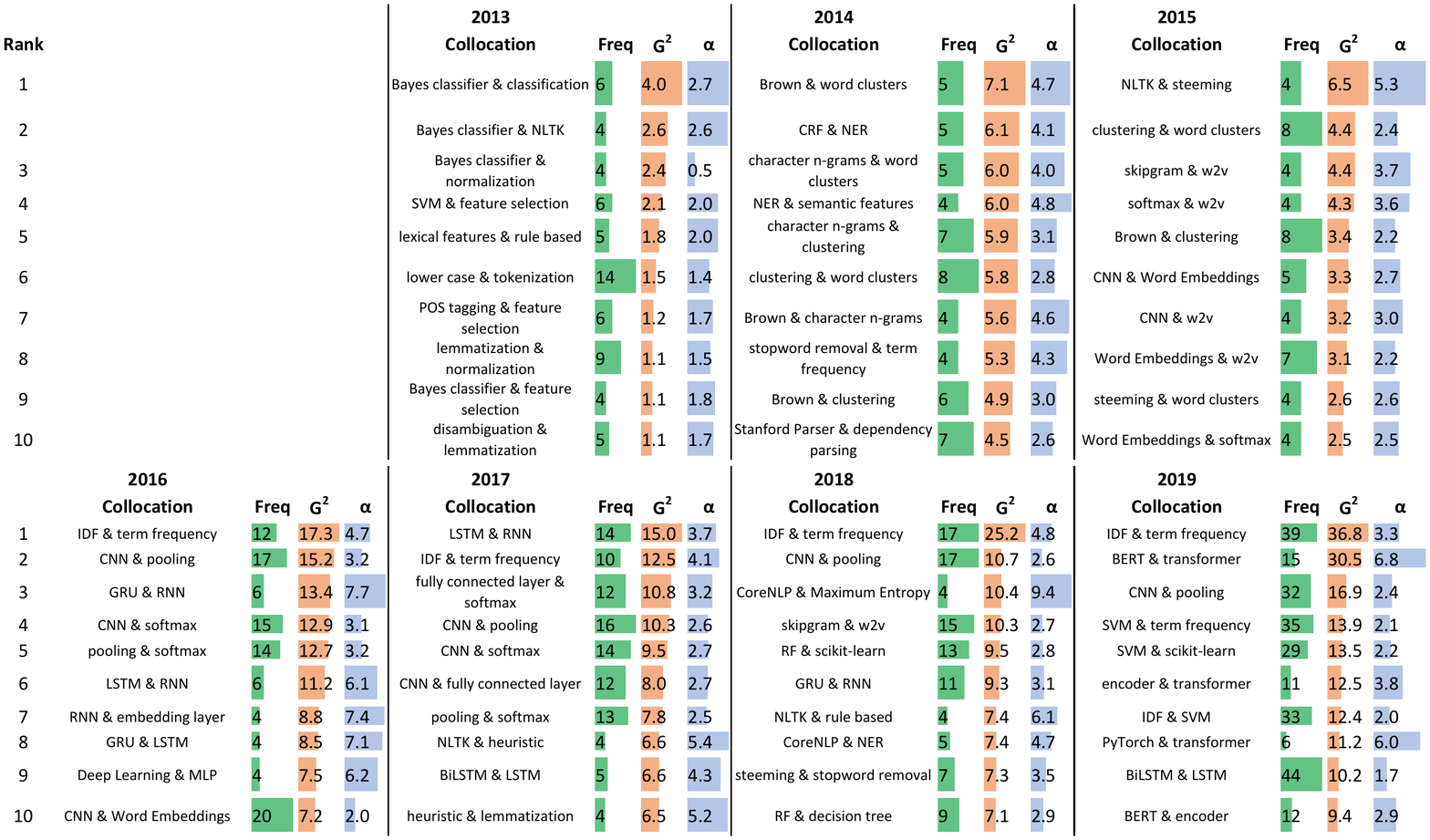}
\caption{Sentiment Analysis tasks in SemEval 2013-2019: Top 10 two-component collocations ranked by $G^2$ (only positive association: odds ratio $\alpha>1$)}
\label{fig:Supp_fig1}
\end{suppfigure}

\begin{suppfigure}%[t!]
\centering
\includegraphics[width= 1\textwidth]{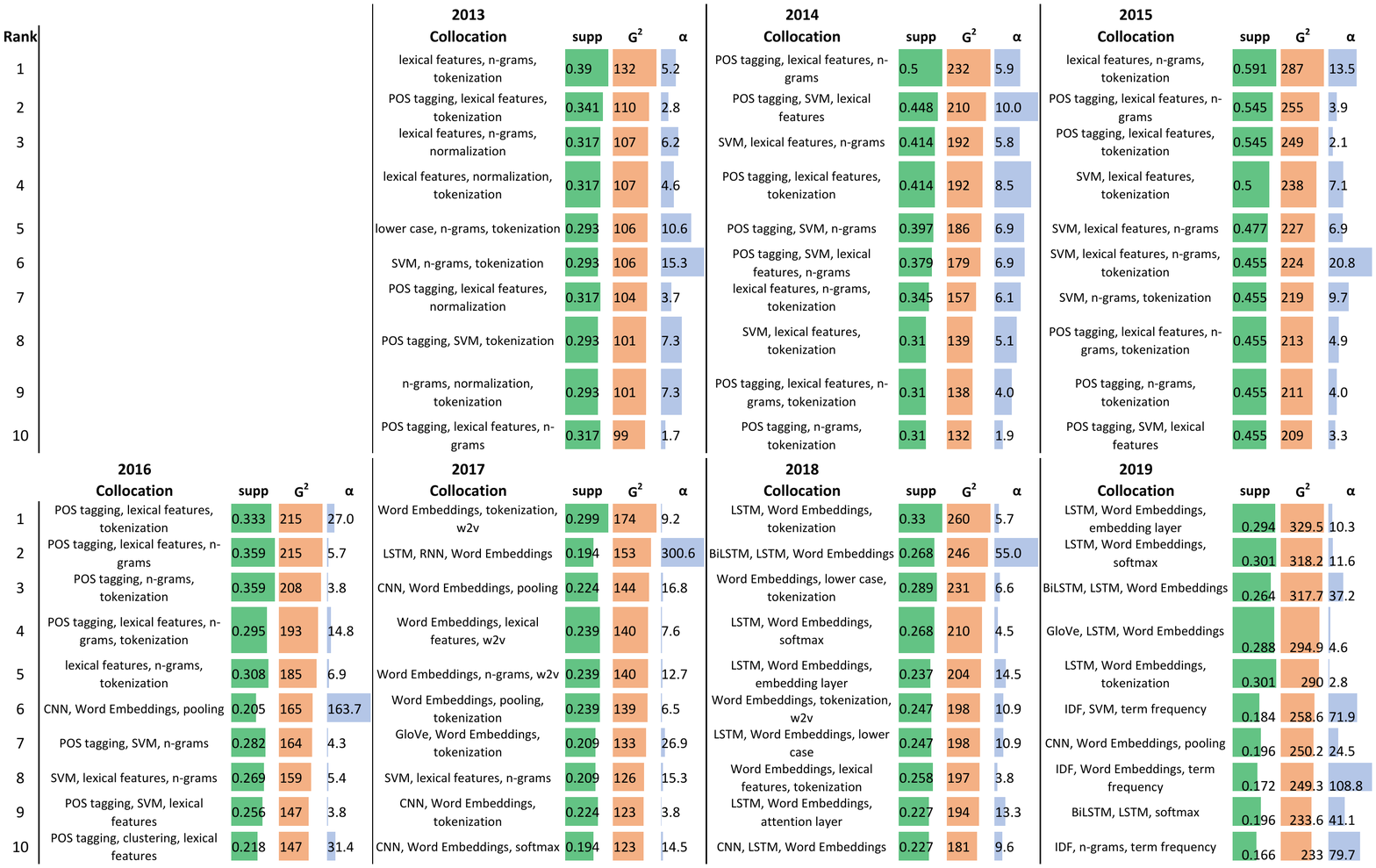}
\caption{Sentiment Analysis tasks in SemEval 2013-2019: Top 10 collocations of $\geq$ 3-components derived from rules found using Apriori algorithm, ranked by $G^2$ (only positive association: odds ratio $\alpha>1$)}
\label{fig:Supp_fig2}
\end{suppfigure}

\begin{suppfigure}%[t!]
\centering
\includegraphics[width= 1\textwidth]{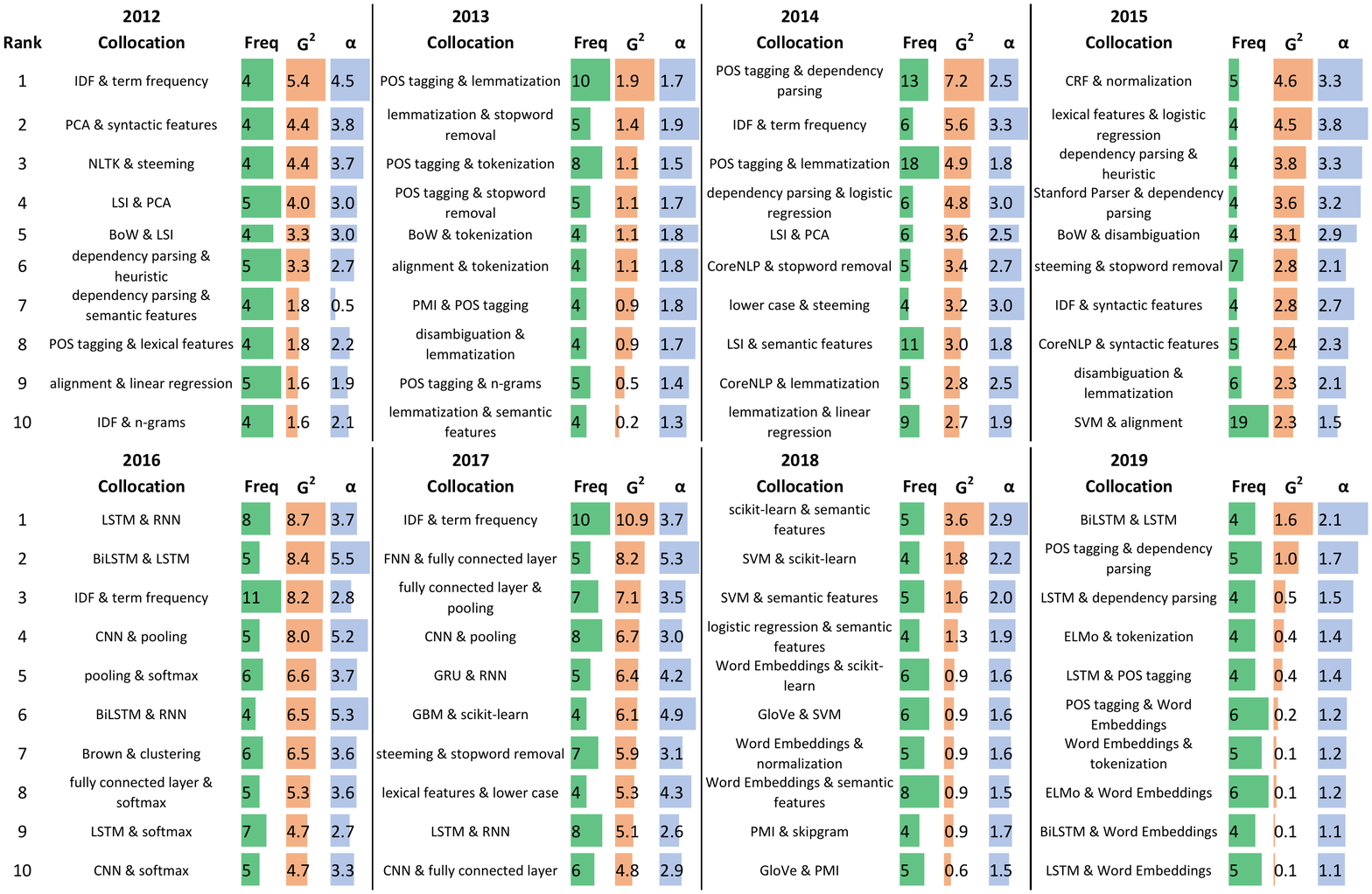}
\caption{Semantics Analysis tasks in SemEval 2013-2019: Top 10 two-component collocations ranked by $G^2$ (only positive association: odds ratio $\alpha>1$)}
\label{fig:Supp_fig3}
\end{suppfigure}

\begin{suppfigure}%[t!]
\centering
\includegraphics[width= 1\textwidth]{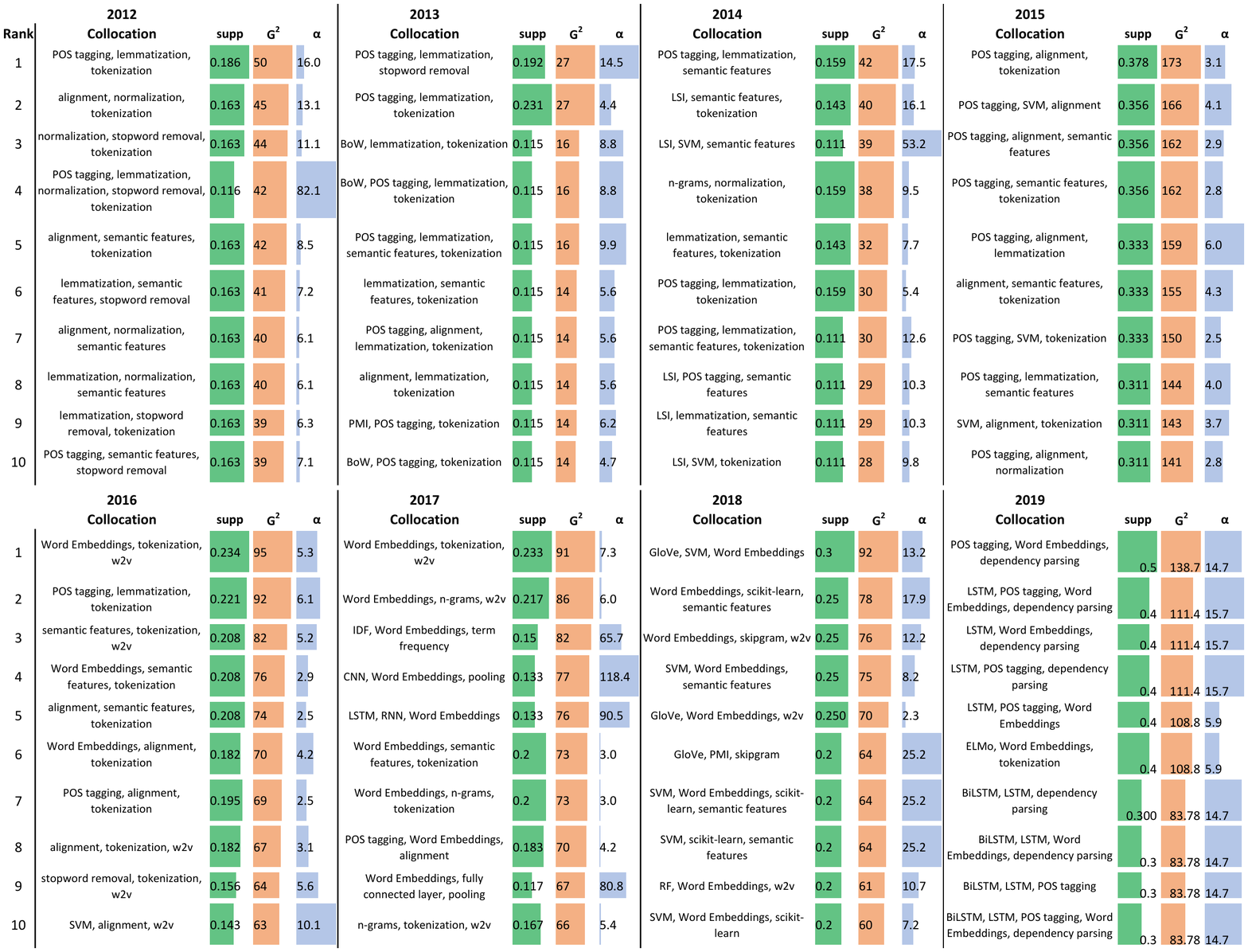}
\caption{Semantics Analysis tasks in SemEval 2013-2019: Top 10 collocations of $\geq$ 3-components derived from rules found using Apriori algorithm, ranked by $G^2$ (only positive association: odds ratio $\alpha>1$)}
\label{fig:Supp_fig4}
\end{suppfigure}

\begin{suppfigure}%[t!]
\centering
\includegraphics[width= 1\textwidth]{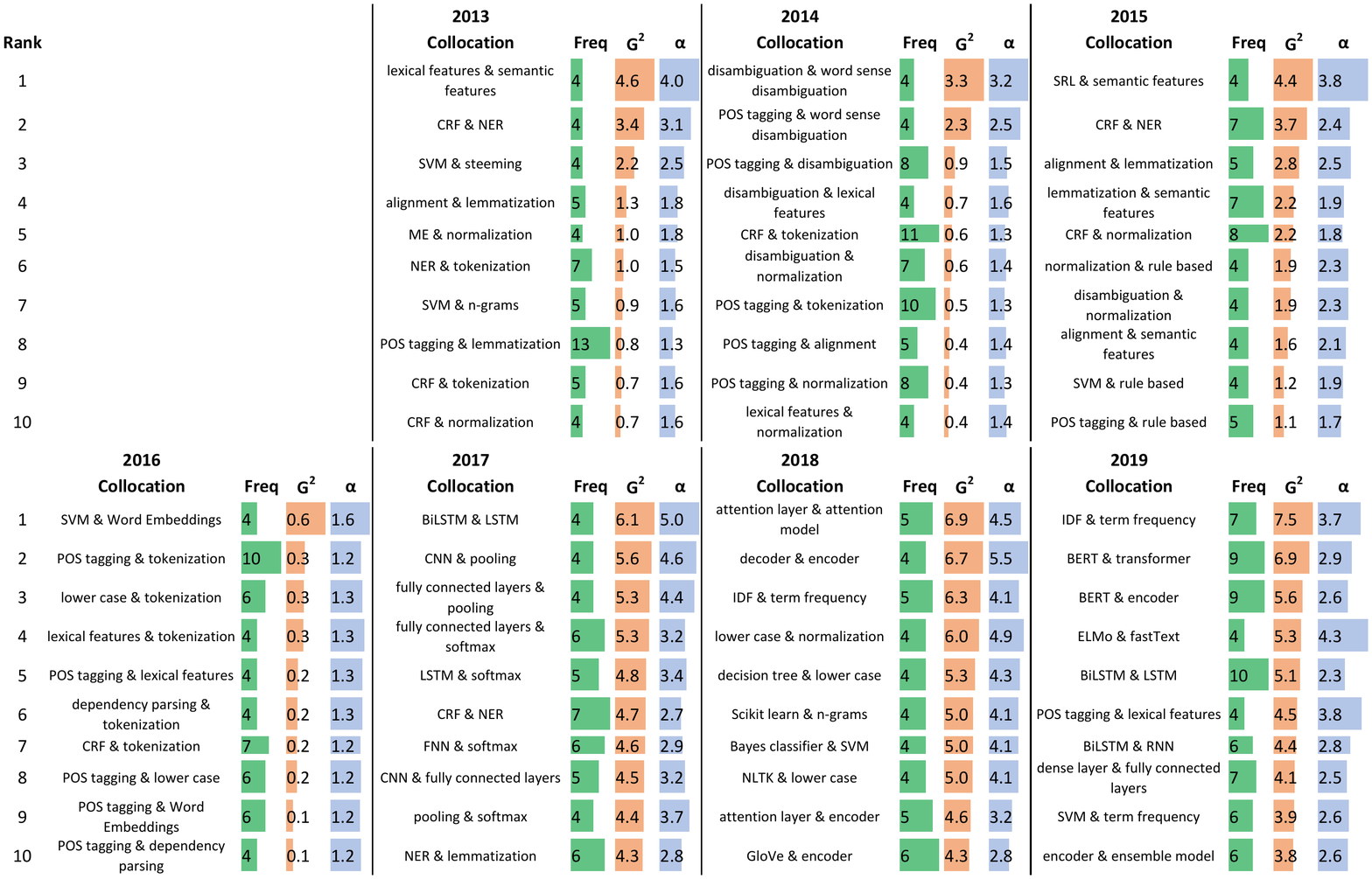}
\caption{Information Extraction tasks in SemEval 2013-2019: Top 10 two-component collocations ranked by $G^2$ (only positive association: odds ratio $\alpha>1$)}
\label{fig:Supp_fig5}
\end{suppfigure}

\begin{suppfigure}%[t!]
\centering
\includegraphics[width= 1\textwidth]{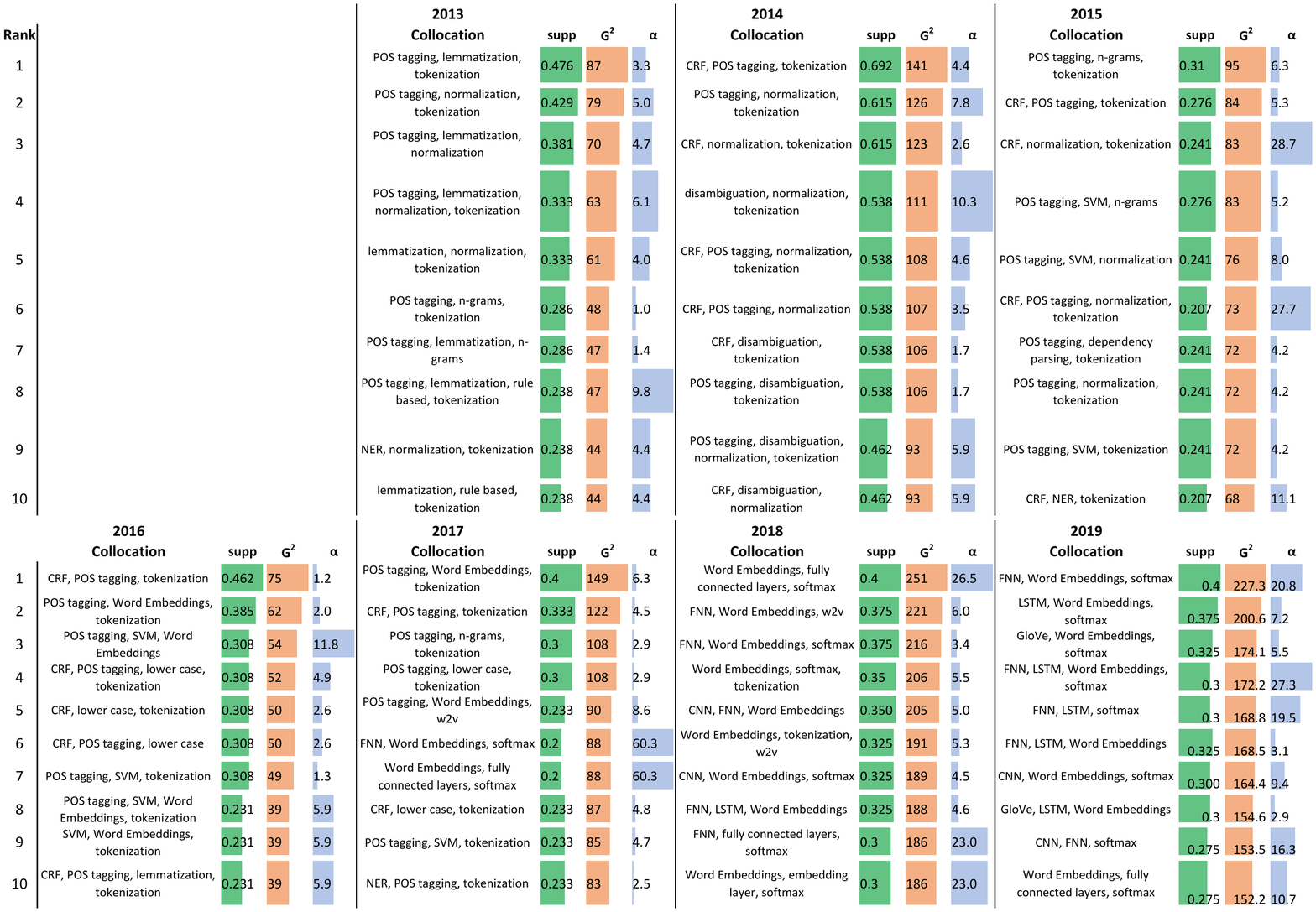}
\caption{Information Extraction tasks in SemEval 2013-2019: Top 10 collocations of $\geq$ 3-components derived from rules found using Apriori algorithm, ranked by $G^2$ (only positive association: odds ratio $\alpha>1$)}
\label{fig:Supp_fig6}
\end{suppfigure}

\begin{suppfigure}%[t!]
\centering
\thisfloatpagestyle{empty}
\includegraphics[width= 1\textwidth]{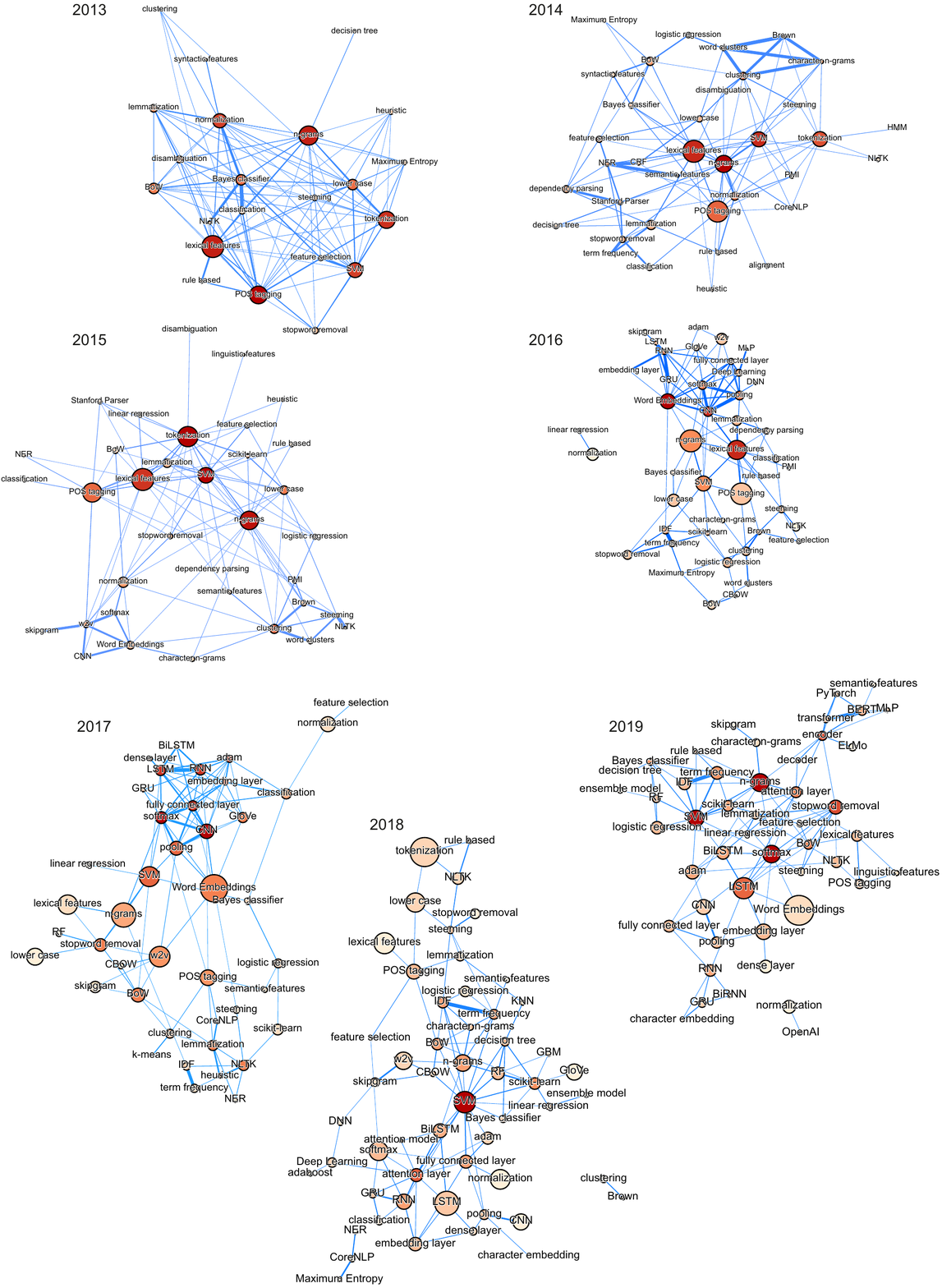}
\caption{Network graphs built using the top 100 collocations from SA tasks in SemEval from 2013 to 2019. Node size is proportional to the frequency of the component, the red color intensity to node degree, edge width to $G^2$ of the collocation. For full resolution see online version of the paper.}
\label{fig:Supp_fig7}
\end{suppfigure}

\begin{suppfigure}%[t!]
\thisfloatpagestyle{empty}
\centering
\includegraphics[width= 1\textwidth]{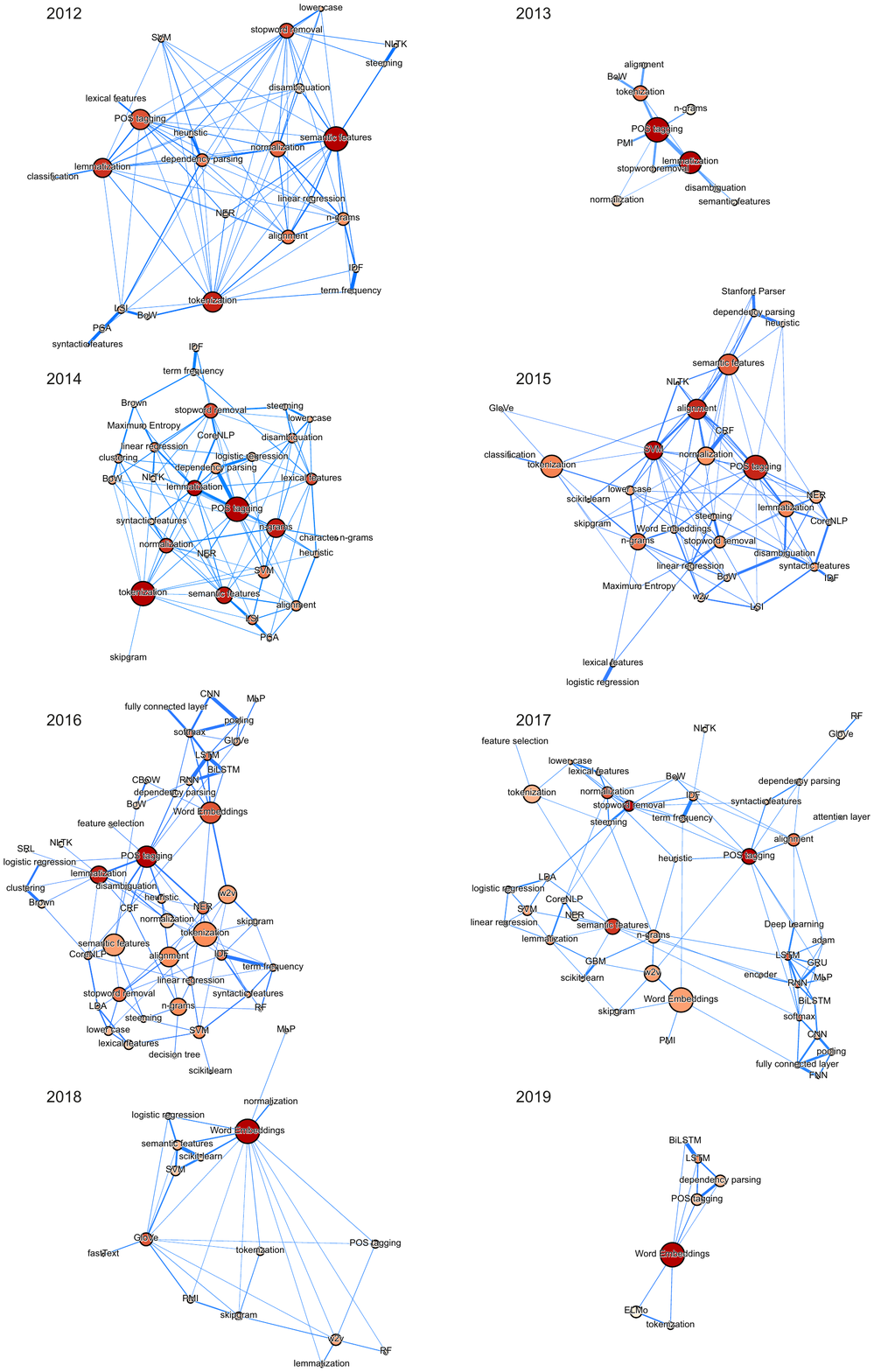}
\caption{Network graphs built using the top 100 collocations from SEM tasks in SemEval from 2012 to 2019. Node size is proportional to the frequency of the component, the red color intensity to node degree, edge width to $G^2$ of the collocation. For full resolution see online version of the paper.}
\label{fig:Supp_fig8}
\end{suppfigure}

\begin{suppfigure}%[t!]
\thisfloatpagestyle{empty}
\centering
\includegraphics[width= 1\textwidth]{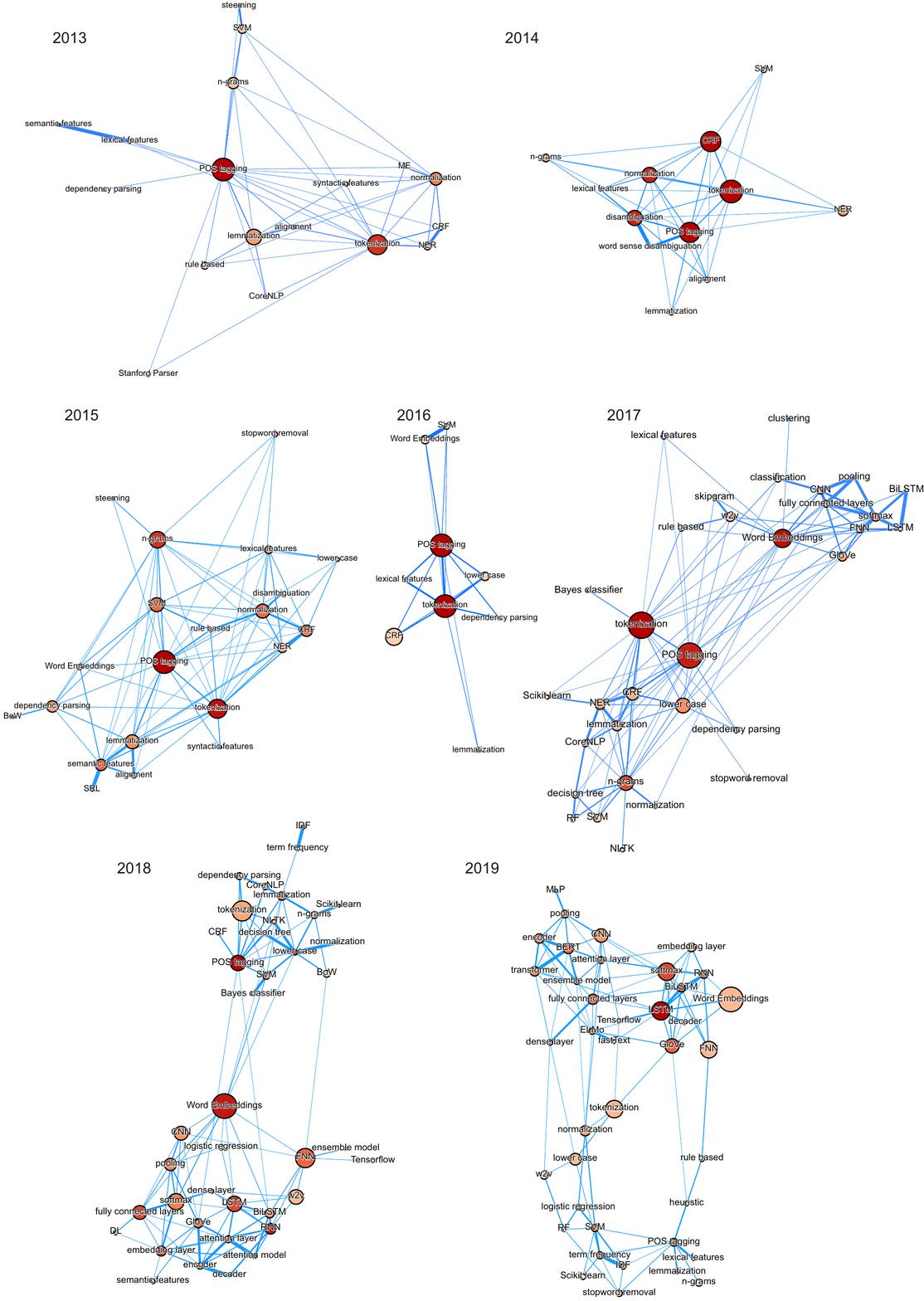}
\caption{Network graphs built using the top 100 collocations from IE tasks in SemEval from 2013 to 2019. Node size is proportional to the frequency of the component, the red color intensity to node degree, edge width to $G^2$ of the collocation. For full resolution see online version of the paper.}
\label{fig:Supp_fig9}
\end{suppfigure}
\clearpage

\end{document}